\documentclass{article}

\usepackage[preprint]{neurips_2025}

\usepackage[dvipsnames,table]{xcolor}
\usepackage{colortbl}

\definecolor{winGreen}{HTML}{009900}
\definecolor{lossRed}{HTML}{CC0000}
\definecolor{boxborder}{HTML}{2E4A45} 
\definecolor{boxbg}{HTML}{F8FAF8}     

\usepackage{graphicx}
\usepackage{booktabs}
\usepackage{multirow}
\usepackage{diagbox}
\usepackage{subcaption}

\usepackage{tcolorbox}
\usepackage{wrapfig}
\usepackage{pifont}       
\usepackage{xspace}
\usepackage{titletoc}
\usepackage{bm}

\usepackage{eccvabbrv}

\usepackage{algorithm}
\usepackage{algorithmic}
\usepackage{arydshln}
\usepackage{dashrule}

\usepackage[accsupp]{axessibility}

\usepackage[colorlinks=true, urlcolor=teal, citecolor=blue, linkcolor=red]{hyperref}
\usepackage{cleveref}

\usepackage{orcidlink}

\newcommand{\methodName}{Immune2V\xspace}

\begin{document}

\noindent
\begin{tcolorbox}[colback=boxbg, colframe=boxborder, arc=3mm, boxrule=1pt, top=6mm, bottom=6mm, left=6mm, right=6mm]

\begin{center}
    {\LARGE \bf \textsf{Immune2V: Image Immunization Against Dual-Stream Image-to-Video Generation} \par}

        \vspace{6mm}
    {\large \textbf{Zeqian Long}$^{*1}$, \textbf{Ozgur Kara}$^{*1}$, \textbf{Haotian Xue}$^{*2}$, \\ \textbf{Yongxin Chen}$^2$, \textbf{James M. Rehg}$^1$ \par}
    \vspace{2mm}
    {\normalsize $^1$University of Illinois Urbana-Champaign, $^2$Georgia Institute of Technology \par}
    {\normalsize $^*$Equal contribution \par}
\end{center}

\noindent\textcolor{gray}{\rule{\linewidth}{0.4pt}}

\noindent \textbf{Abstract.} Image-to-video (I2V) generation has the potential for societal harm because it enables the unauthorized animation of static images to create realistic deepfakes. While existing defenses effectively protect against static image manipulation, extending these to I2V generation remains underexplored and non-trivial. In this paper, we systematically analyze why modern I2V models are highly robust against naive image-level adversarial attacks (\ie, immunization). We observe that the video encoding process rapidly dilutes the adversarial noise across future frames, and the continuous text-conditioned guidance actively overrides the intended disruptive effect of the immunization. Building on these findings, we propose the \textbf{Immune2V} framework which enforces temporally balanced latent divergence at the encoder level to prevent signal dilution, and aligns intermediate generative representations with a precomputed collapse-inducing trajectory to counteract the text-guidance override. Extensive experiments demonstrate that Immune2V produces substantially stronger and more persistent degradation than adapted image-level baselines under the same imperceptibility budget.
\vspace{6mm}

\noindent \textbf{GitHub:} \url{https://github.com/Zeqian-Long/Immune2V} \\
\textbf{Project Webpage:} \url{https://immune2v.github.io/} \\

\end{tcolorbox}

\begin{center}
    \captionsetup{type=figure}
\includegraphics[width=\linewidth]{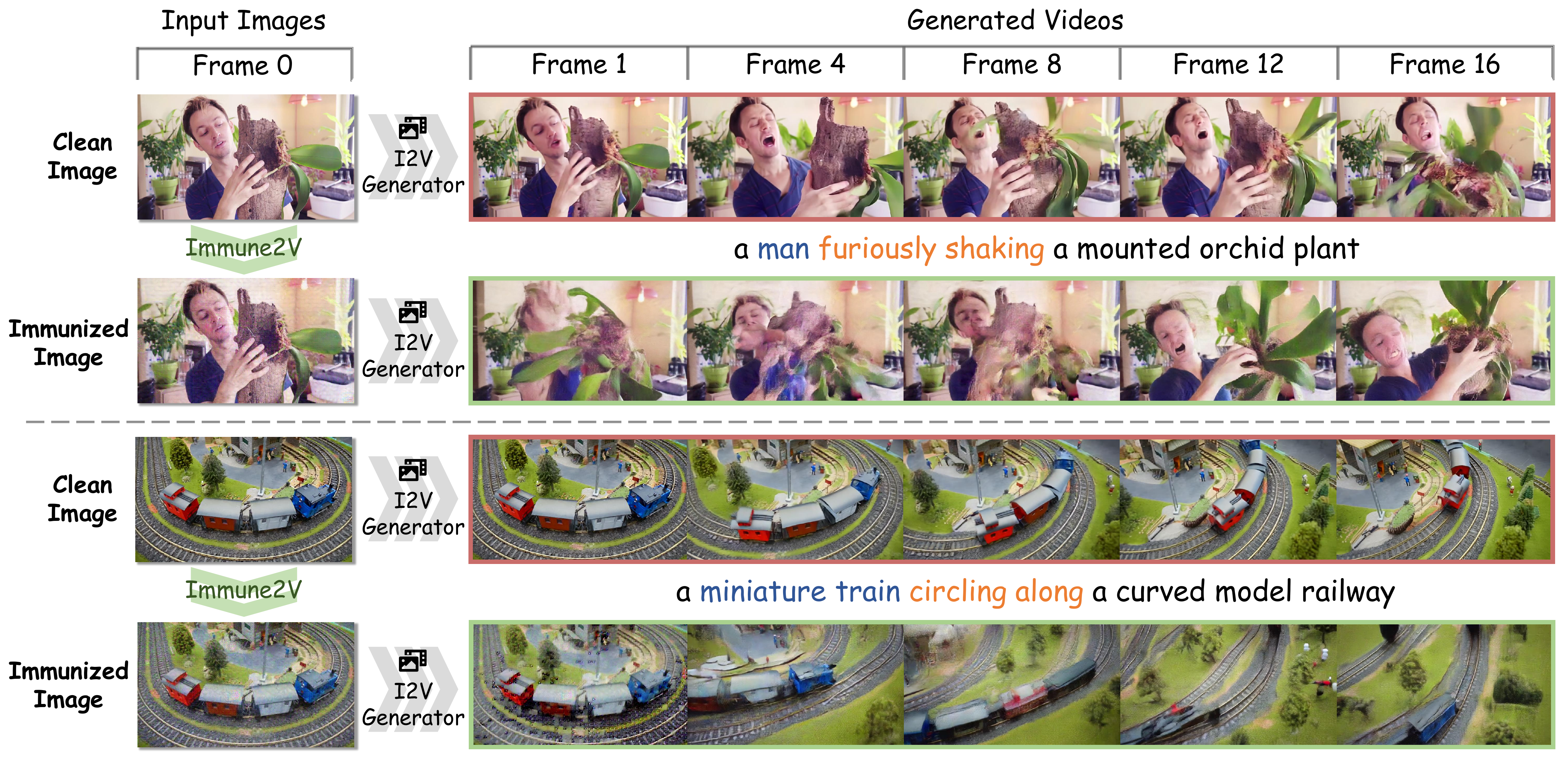}
\caption{\textit{\textbf{Immune2V Protection.}} \methodName protects images from Image-to-Video (I2V) generation by adding imperceptible immunization noise to produce an immunized image. When processed by an I2V generator, a clean image yields realistic motion (\eg, the human face articulates naturally and the train moves circularly along the track, top row), whereas the immunized image disrupts the generation dynamics, producing highly unrealistic videos (\eg, the human subject is severely distorted and the train structurally collapses, failing to synthesize coherent motion, bottom row).
    }
    \label{fig:teaser}
\end{center}

\section{Introduction}
\label{sec:intro}







\begin{figure}[t]
  \centering
  \includegraphics[width=\linewidth]{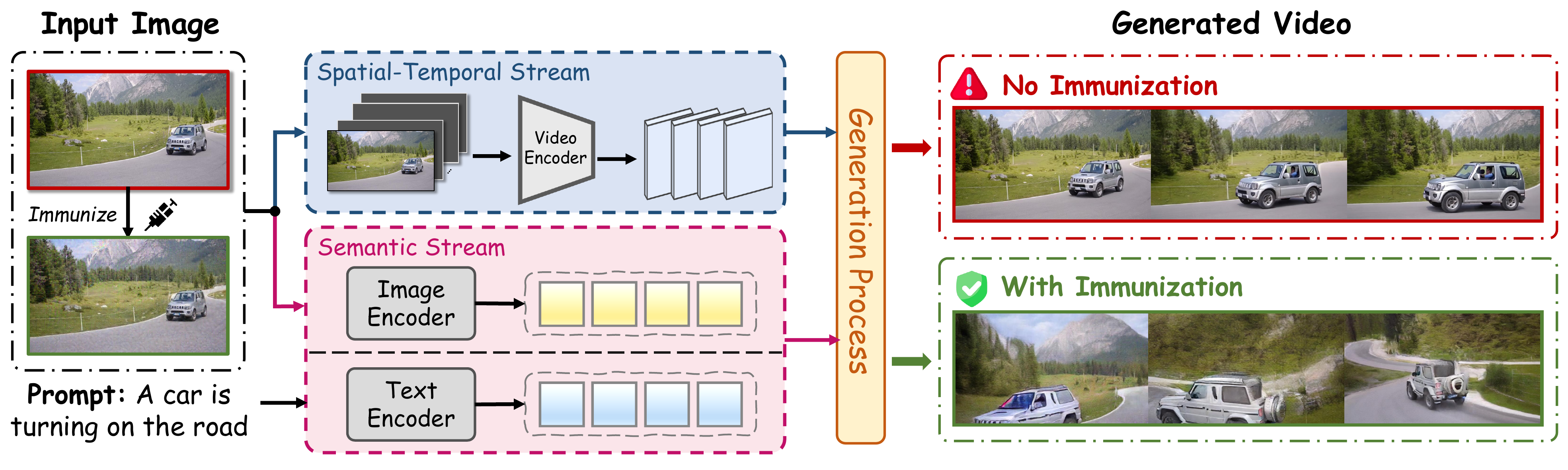}
  \caption{\textit{\textbf{Immunization on Dual-Stream I2V Architecture.}} A \textcolor[HTML]{c00000}{clean input image}, guides the generation process through i) a \textbf{spatial-temporal stream}, where the image is processed through a video encoder to initialize the structural latent space, and ii) a \textbf{semantic stream}, where an image encoder extracts high-level embeddings for continuous semantic guidance. While these streams normally generate a realistic video conditioned on an input image, injecting an adversarial perturbation yields an \textcolor[HTML]{548235}{immunized image} that simultaneously disrupts both conditioning streams, forcing the generation process into structural collapse.}
  \label{fig:dual}
\end{figure}

While the rapid advancement of generative models~\cite{ho2020denoising,blattmann2023stable,ho2022imagen, liu2024sora} enables the synthesis of highly realistic content, it also introduces the potential for substantial harm by allowing malicious actors to create convincing deepfakes~\cite{masood2021deepfakes}. A particularly concerning threat is Image-to-Video (I2V) generation, which enables the unauthorized animation of static personal images harvested online into fabricated videos via semantic conditioning (\eg text prompts)~\cite{masood2021deepfakes, wan2025wanopenadvancedlargescale, liu2024sora}. Although significant efforts have been made in deepfake detection~\cite{rana2022deepfake}, these methods are inherently reactive. An alternative proactive approach uses adversarial perturbation methods in discriminative vision tasks~\cite{goodfellow2014explaining, kurakin2018adversarial, arnab2018robustness, jia2025adversarial} and image diffusion models~\cite{liang2023adversarial, salman2023raising, liang2023mist, xue2024effectiveprotectiondiffusionbased, lo2024distraction, liu2024metacloak, glaze, ahn2024imperceptible, ozden2026diffvax}. This approach adopts the input-level strategy of \emph{immunization}, which injects an optimized perturbation directly into the image before it is shared online. Ideally, if an adversary subsequently attempts to animate such an immunized image, the embedded perturbation would disrupt the generative model's internal representations, causing the synthesized video to become severely distorted (Fig. \ref{fig:dual}).


Based on the success of image immunization, a natural question is: 
\textit{\textbf{Why not naively extend existing immunization methods to the video domain?}} While this might appear straightforward, the I2V setting introduces fundamental challenges. First, because I2V models receive strong conditioning via the input image that anchors content generation, an attack cannot simply drive the output toward trivial or unrelated results (\eg, blank frames), but must instead induce semantic and/or structural failures under the constraint of the conditioning signal. Second, analysis of the  architectures of modern I2V models reveals an even deeper challenge: State-of-the-art systems adopt a ``dual-stream'' image conditioning architecture (Fig.~\ref{fig:dual}), in which a single input image independently provides both spatial-temporal and semantic conditioning. As we show in Sec.~\ref{sec:fail}, this creates two bottlenecks that neutralize naive image-level attacks. In the spatial-temporal conditioning stream (Fig.~\ref{fig:dual}, blue box), the input image is mapped into a sequence of latents by a video encoder, which in many implementations takes the form of a 3D VAE. Causality is introduced in most 3D VAEs to enable more efficient inference by reducing the temporal footprint. This design causes perturbations on the first frame to attenuate rapidly, since long-range dependencies diminish quickly in the causal decoder.

In addition, the semantic conditioning stream (Fig.~\ref{fig:dual}, pink box) introduces a second layer of robustness via text guidance. In this stream, semantic image latents produced by an image encoder are paired with semantic embeddings from the text prompt and injected to guide the generation process for each video frame. We show that when prompts are \emph{aligned} with the initial image (\ie, the text prompt matches the image content), the semantic guidance is reinforced and demonstrates
a strong ``healing'' effect--- even if the spatial-temporal stream is compromised via naive immunization, the continuous semantic pull of a coherent prompt actively overrides the visual corruption and forces the trajectory back to a consistent structure. As a consequence, \emph{the combination of rapid temporal attenuation and continuous text conditioning renders standard image attacks completely ineffective in achieving persistent disruption of video generation.}

To address this challenge, we introduce \textbf{\methodName}, a dual-stream joint optimization framework that explicitly targets both the spatial and semantic streams of I2V models. First, to address the rapid gradient attenuation in the spatial stream, we introduce a \textbf{Spatial-Temporal Attack}, which enforces temporally-balanced supervision directly at the encoder level. By aligning the corrupted latents with a target video across all temporal segments, we prevent the immunization signal from vanishing, forcing the adversarial effect to persist from the first frame to the last. Second, to counter the continuous conditioning override in the semantic stream, we design a \textbf{Semantic Attack}, which utilizes the trajectory of latents generated during sampling via a mismatched, or ``bad'' prompt, as a proxy. By precomputing this degraded trajectory and treating it as a pseudo ground truth, we optimize the adversarial image so its internal semantic features actively push the generation process toward this target. Consequently, during inference time, when an attacker provides a normal prompt, our method hijacks the iterative guidance and forces the video into a state of structural breakdown. By jointly optimizing these two attacks into a single immunization, \methodName successfully bypasses the dual-stream constraints, inducing severe and long-lasting degradation across the entire generated video.

In validating the benefit of Immune2V, we establish comprehensive quantitative metrics for video quality and consistency, and introduce Vision Language Models (VLMs) as automated judges for qualitative evaluation. Our main contributions are:

\begin{itemize}
    \item A systematic analysis of why a naive extension of image-level immunization fails in I2V generation, identifying temporal attenuation in the spatial stream and conditioning override in the semantic stream as the two key bottlenecks.
    
    \item A joint dual-stream optimization framework \methodName that simultaneously targets spatial and semantic conditioning, enabling a single immunization to induce persistent degradation across the generated video.
    
    \item A comprehensive video-level evaluation protocol, including structural consistency metrics and VLM-based automatic judgment, to rigorously assess immunization effectiveness.
\end{itemize}

\section{Related Work}
\label{sec:related}

\subsection{Image-To-Video Generation}
Diffusion- \cite{ho2020denoising, singer2023makeavideo, blattmann2023align, bar2024lumiere, ho2022imagen} and flow-based \cite{liu2022flow, wan2025wanopenadvancedlargescale, chen2025goku, jin2024pyramidal} text-to-video (T2V) models synthesize high-quality videos. Building upon these, image-based conditioning guides synthesis via semantic guidance \cite{xing2024dynamicrafter, zhang2023i2vgen}, visual consistency \cite{ren2024consistiv, hu2024animate}, and explicit structural control \cite{yu2023animatezero, guo2023animatediff, peng2024controlnext}. 
By leveraging complementary information from image and text inputs, these conditioning mechanisms enable generation to achieve both high visual quality and improved controllability \cite{long2026followyourshape, ma2025controllable, ma2025follow}.
Modern state-of-the-art models—such as Wan \cite{wan2025wanopenadvancedlargescale}, I2VGen-XL \cite{zhang2023i2vgen}, and DynamiCrafter \cite{xing2024dynamicrafter}—predominantly employ a dual-stream architecture for scalable video generation \cite{blattmann2023stable, yang2024cogvideox, kong2024hunyuanvideo, kling2024, luma2024dreammachine}. Earlier diffusion approaches (\eg, I2VGen-XL, DynamiCrafter) rely on explicit image conditioning to guide motion synthesis and maintain visual consistency. To improve scalability and generalization, Wan \cite{wan2025wanopenadvancedlargescale} adopts a rectified-flow formulation and large-scale unified pre-training for strong performance across diverse I2V scenarios. Despite these advances, the impact of input-level perturbations on such dual-stream pipelines remains largely underexplored.

\subsection{Adversarial Attacks on Generative Diffusion Models}
Adversarial attacks have been extensively studied in diffusion-based systems (\eg, image generation \cite{salman2023raising, liang2023adversarial, liang2023mist, shen2025decontext, xue2024effectiveprotectiondiffusionbased, xue2024pixel, liu2024metacloak, ahn2024imperceptible}, video generation~\cite{gui2025i2vguard, li2024prime, li2025t2vattack} and policy generation~\cite{chen2024diffusion, kalra2025vulnerable}). For image diffusion, methods like PhotoGuard \cite{salman2023raising}, DiffVax \cite{ozden2026diffvax}, Mist~\cite{liang2023mist} and SDS-Attack~\cite{xue2024effectiveprotectiondiffusionbased} design perturbations to disrupt encoders and diffusion dynamics, while DAYN \cite{lo2024distraction} and DiffusionGuard \cite{choi2024diffusionguard} target cross-attention and localized masks. However, extending this success to video diffusion is non-trivial. Existing studies remain limited: they either focus on jailbreaking text prompts~\cite{li2025t2vattack} or target outdated models like SVD~\cite{blattmann2023stable}, which heavily inherit from image diffusion. Recent work like I2VGuard~\cite{gui2025i2vguard} explores I2V immunization; however, it shows clear improvements only on SVD, yielding marginal gains on recent models like CogVideo-X~\cite{yang2024cogvideox}, which they attribute to the robustness of modern 3D VAEs.

\section{Protection Against I2V Generation via Immune2V}

\subsection{Problem Setting: Input-Image Immunization for I2V Models}

Image-to-Video (I2V) generation synthesizes a video $\mathcal{V} = G(\cdot\,; I, c)$, using a generative model $G$ conditioned on a clean static image $I \in \mathbb{R}^{C \times H \times W}$ and a semantic condition $c$ (\eg text prompt). To protect this content against unauthorized generation, input-level ``immunization'' embeds an imperceptible adversarial perturbation $\delta \in \mathbb{R}^{C \times H \times W}$ directly into the source image, yielding an immunized image $I_{\mathrm{adv}} = I + \delta$. This embedded noise is specifically crafted to disrupt the model's generative process. We achieve this by optimizing the perturbation to minimize the distance between the video generated from the immunized image and a predefined, severely degraded target video $\mathcal{V}_{\mathrm{target}}$ (\eg a completely zeroed video). This defense is formulated as the following objective:
\begin{equation}
\label{eq:standard}
\delta^\ast = \arg\min_{\lVert \delta \rVert_\infty \leq \epsilon}
\mathcal{L}\big(G(\cdot,; I_{\mathrm{adv}}, c),; \mathcal{V}_{\mathrm{target}}\big).
\end{equation}
where the loss function $\mathcal{L}$ quantifies the distance between the synthesized output and the reference target $\mathcal{V}_{\mathrm{target}}$.
A successful defense must induce severe and persistent degradation across the entire generated video. The core challenge in the I2V setting is ensuring that this protective effect does not merely corrupt the initial frames, but actively propagates throughout the full temporal length of the sequence.


\subsection{Why Standard Image Attacks Fail on Dual-Stream I2V Models?}
\label{sec:fail}

In this section, we analyze why standard adversarial attacks (Eq.~\ref{eq:standard}), despite their success in image diffusion~\cite{liang2023adversarial, xue2024effectiveprotectiondiffusionbased, liang2023mist}, fail to reliably disrupt I2V models. 
We specifically focus on the \textit{dual-stream} conditioning paradigm, which has become a dominant design in many state-of-the-art (SOTA) I2V models (\eg, Wan~\cite{wan2025wanopenadvancedlargescale}, I2VGen-XL~\cite{zhang2023i2vgen}, and DynamiCrafter~\cite{xing2024dynamicrafter}).
Such models typically combine i) structural initialization from the input image and ii) semantic guidance from image and text embeddings during denoising.
We use Wan \cite{wan2025wanopenadvancedlargescale} as recent large-scale and representative instantiation of this paradigm. In this model, the input image $I$ and text prompt $c$ influence the model through two streams: i) a \textbf{spatial-temporal stream}, where the image is temporally expanded into a zero-padded pseudo-video $\tilde{I} = [I, 0, \ldots, 0]$. 
A 3D Causal VAE encoder $\mathcal{E}_{\text{enc}}$ then maps it into a spatio-temporal latent tensor $z = \mathcal{E}_{\text{enc}}(\tilde{I}) \in \mathbb{R}^{C' \times T' \times H' \times W'}$, which initializes and structurally conditions the denoising process; and ii) a \textbf{semantic stream}, where high-level embeddings $e_{\text{img}} = \mathcal{E}_{\text{img}}(I)$ and $e_{\text{txt}} = \mathcal{E}_{\text{txt}}(c)$ are injected into the diffusion backbone $f_\theta$ via cross-attention to maintain semantic alignment. 
We identify a structural bottleneck within each pathway that inherently limits the temporal propagation and effectiveness of standard image-level perturbations (Fig.~\ref{fig:attenuation} and Fig.~\ref{fig:semantic_override}).


\subsubsection{i) Perturbation Attenuation in the Spatial-Temporal Stream}
\mbox{} 

\begin{tcolorbox}[width=\linewidth, colback=blue!5!white, colframe=blue!75!black, arc=2mm, boxrule=0.5pt, left=4pt, right=4pt, top=4pt, bottom=4pt]
\textbf{Observation 1:} In the spatial-temporal stream, the influence of adversarial noise applied to the input image rapidly vanishes in later frames during both forward and backward propagation.
\end{tcolorbox}

\begin{figure*}[t!]
\centering
\begin{subfigure}{\linewidth}
    \centering
    \includegraphics[width=\linewidth]{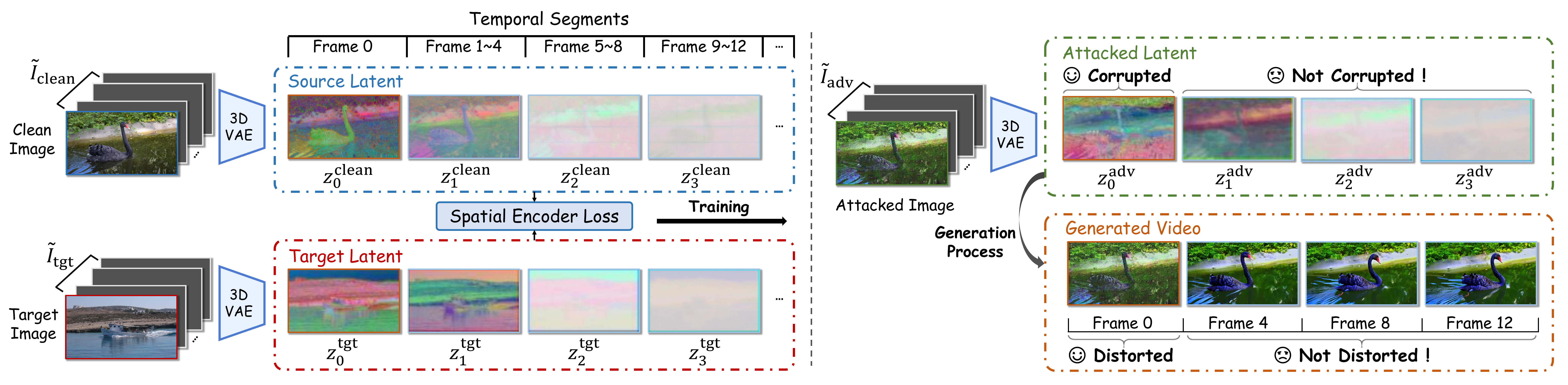}
    \caption{\textit{\textbf{Latent Visualization.}} Naive extension of image-diffusion attacks~\cite{liang2023mist, liang2023adversarial} optimizes the adversarial noise by minimizing the $L_2$ distance between clean ($z^{\mathrm{clean}}$) and target ($z^{\mathrm{tgt}}$) 3D VAE encoded latents. Testing the attacked image ($\tilde{I}_{\mathrm{adv}}$) reveals that only the first frame's latent is effectively corrupted, allowing the generated video to quickly recover its original content in subsequent frames.}
\end{subfigure}
\vspace{0.5em}
\begin{subfigure}{0.46\linewidth}
    \centering
    \includegraphics[width=\linewidth]{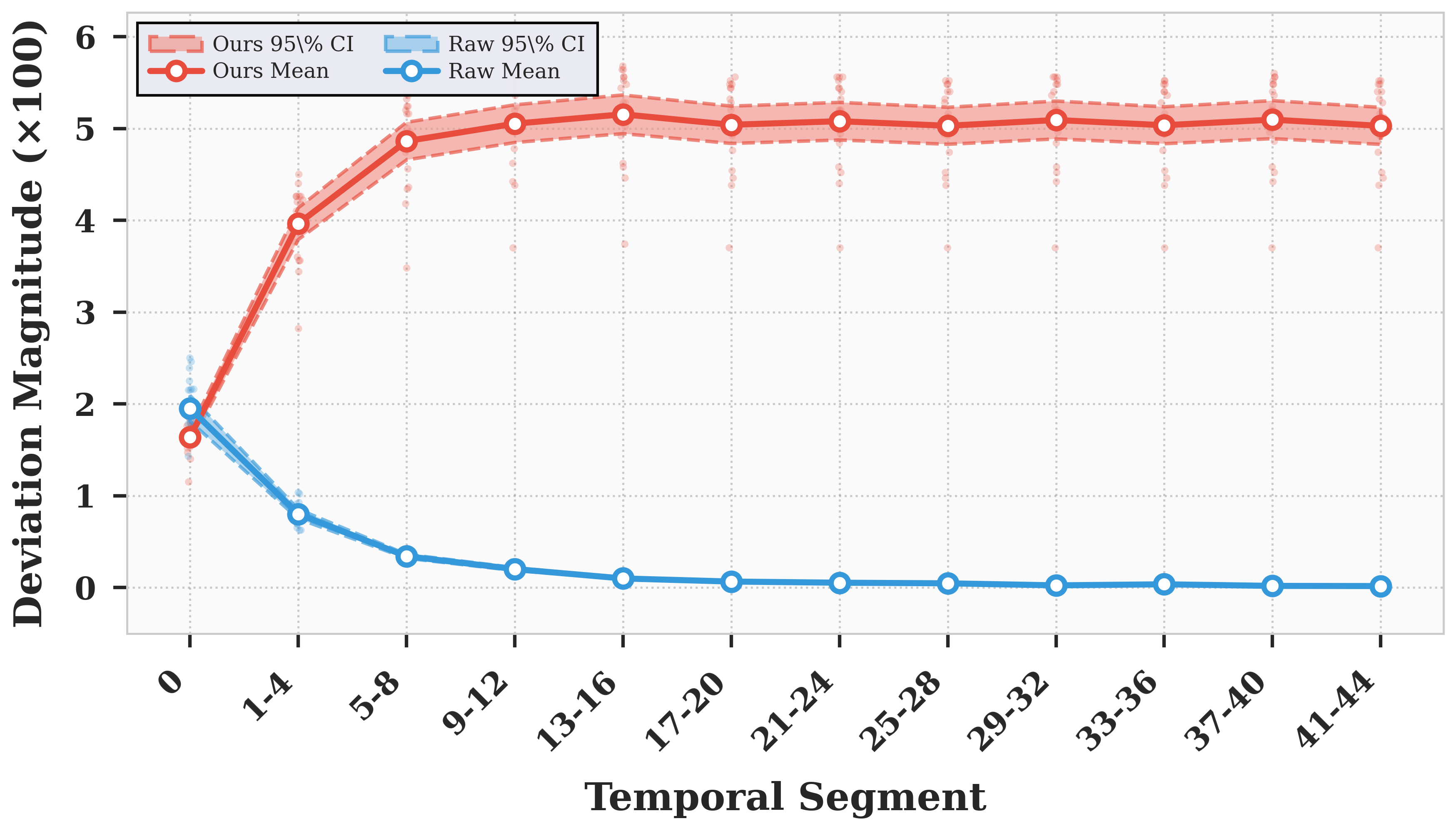}
    \caption{\textit{\textbf{Latent Deviation ($d_t = \|z_t^{\mathrm{tgt}} - z_t^{\mathrm{clean}}\|_2$).}}}
\end{subfigure}
\hfill
\begin{subfigure}{0.46\linewidth}
    \centering
    \includegraphics[width=\linewidth]{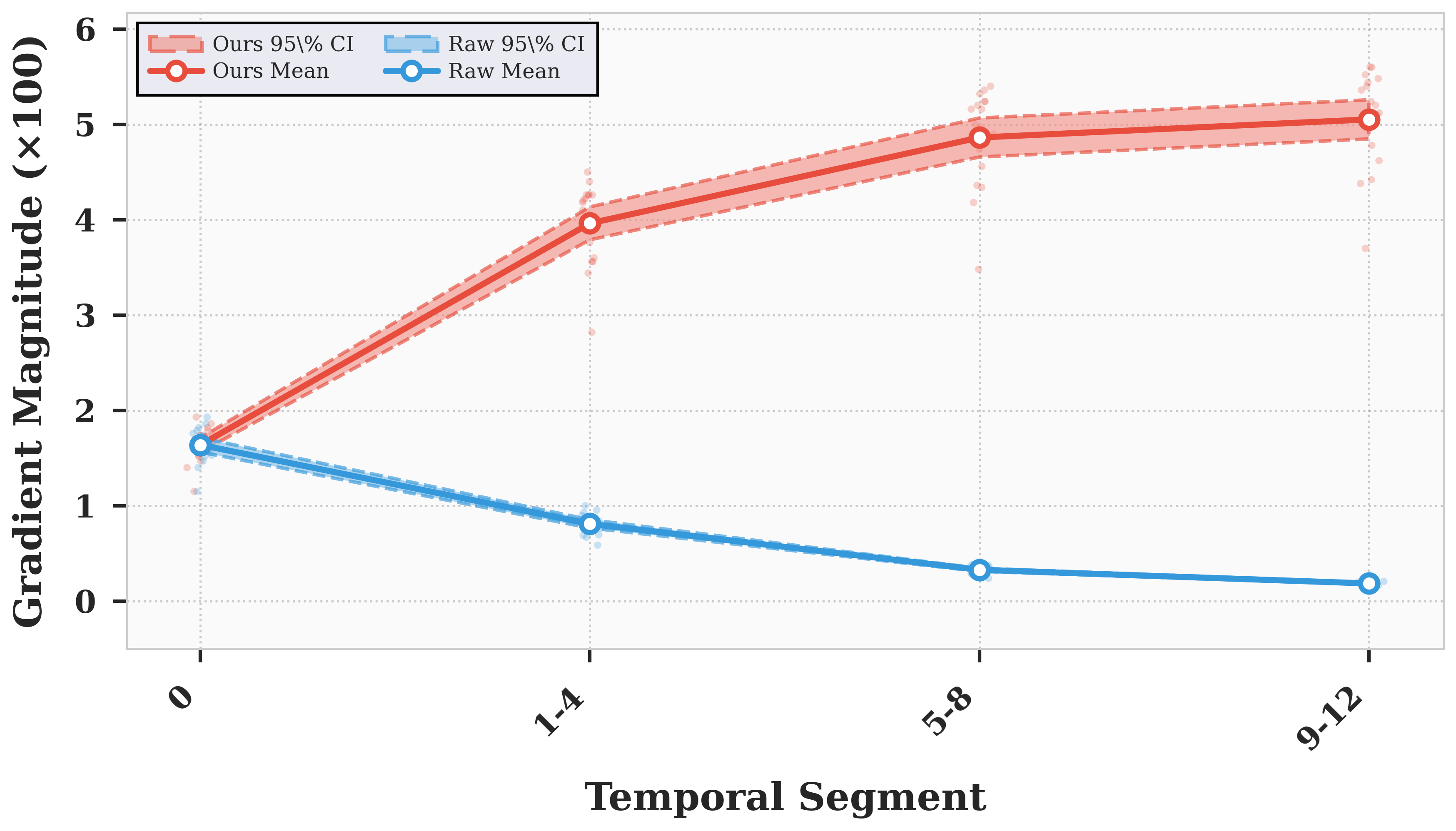}
    \caption{\textit{\textbf{Gradient Magnitude ($g_t = \|\nabla_I d_t\|_2$).}}}
\end{subfigure}
\caption{\textit{\textbf{Temporal Attenuation.}} Standard \textcolor[HTML]{3498DB}{image-level attacks} fail because their influence vanishes rapidly across the zero-padded temporal segments during both forward and backward propagation. \textcolor[HTML]{E74C3C}{Our temporally-balanced approach} overcomes this by actively enforcing persistent corruption across the entire temporal axis. Due to memory constraints, we compute the gradient for frames 0--12.}
\label{fig:attenuation}
\end{figure*}

The spatial-temporal stream operates on the pseudo-video $\tilde{I} = [I,0,\ldots,0]$ with its encoded latent representation $z = \mathcal{E}_{\text{enc}}(\tilde{I})$. We index the latent tensor along the temporal dimension as $z=\{z_t\}_{t=1}^{T'}$, where $z_t\in\mathbb{R}^{C' \times H' \times W'}$ denotes the latent slice corresponding to the $t$-th temporal segment. Due to the causal spatio-temporal convolution structure, each slice depends only on past frames of the input sequence.

Applying the standard adversarial attack (Eq.~\ref{eq:standard}) perturbs the input image as $I+\delta$. Following prior image-diffusion attacks, the perturbation is optimized using a spatial encoder objective that aligns the encoded representations of the clean and target sequences, measured by the latent deviation
\begin{equation}
    d_t = \|z_t^{\mathrm{tgt}} - z_t^{\mathrm{clean}}\|_2.
\end{equation}

However, as shown in Fig.~\ref{fig:attenuation}(a), the adversarial perturbation primarily affects the first temporal segment, while later slices rapidly revert to their clean states, causing the distortion to vanish in subsequent frames of the generated videos. Formally, during the generation process, the perturbed image is expanded into $\tilde{I}_{\mathrm{adv}} = [I+\delta,0,\ldots,0]$. Since this pseudo-video contains only a single non-zero frame, later temporal receptive fields predominantly observe zero-valued inputs. Combined with the temporal downsampling from $T$ input frames to $T'$ latent slices in the causal 3D VAE~\cite{gui2025i2vguard}, the adversarial signal introduced in the first frame rapidly attenuates along the temporal dimension.

To analyze this effect quantitatively, we examine both the forward and backward passes of the spatial-temporal stream. In the forward pass, Fig.~\ref{fig:attenuation}(b) plots the per-slice deviation $d_t$ across the temporal axis. Under the standard attack (blue curve), $d_t$ exhibits a clear monotonic decay as $t$ increases, indicating that the adversarial influence progressively weakens across later latent slices. In the backward pass, we compute the slice-wise gradient magnitude $g_t = \|\nabla_I d_t\|_2$, which measures the influence of each temporal slice on the input image during optimization. 
Fig.~\ref{fig:attenuation}(c) shows that the gradients from later slices ($t \to T'$) become negligible. 
As a result, the optimization process is dominated by the earliest slices, leaving the latent states responsible for later video frames effectively unattacked.

Overcoming this limitation requires an adversarial objective that prevents temporal attenuation and redistributes the optimization signal across the entire sequence. 
Our solution (Sec.~\ref{sec:solution}) is based on a fully-populated target sequence together with slice-wise supervision across the latent timeline. As illustrated by the red curves in Fig.~\ref{fig:attenuation}(b) and (c), this formulation maintains strong latent deviation and balanced gradient signals across all temporal slices.

\subsubsection{ii) Conditioning Override in the Semantic Stream}
\mbox{} 

\begin{tcolorbox}[width=\linewidth, colback=blue!5!white, colframe=blue!75!black, arc=2mm, boxrule=0.5pt, left=4pt, right=4pt, top=4pt, bottom=4pt]
\textbf{Observation 2:} In the semantic stream, text-conditioned guidance dominates the generative trajectory. Because of this dominance, an aligned prompt (good prompt) can effectively override perturbations arising from spatial-only attacks, whereas a mismatching prompt (bad prompt) drives the generation toward unreasonable and divergent outputs.
\end{tcolorbox}


In the semantic stream, the image condition and text prompt are injected through cross-attention layers in the transformer blocks. The attention output is the sum of the text- and image-conditioned responses:
\begin{equation}
    \mathbf{F}_{\mathrm{out}} = \mathbf{F}_{\mathrm{text}} + \mathbf{F}_{\mathrm{img}}
\end{equation}
where $\mathbf{F}_{\mathrm{text}}$ and $\mathbf{F}_{\mathrm{img}}$ are the corresponding attention outputs for the text and image. When $\mathbf{F}_{\mathrm{text}}$ is misaligned with $\mathbf{F}_{\mathrm{img}}$, their conflicting signals cause the combined output $\mathbf{F}_{\mathrm{out}}$ to diverge. Because this cross-attention mechanism exists in every transformer layer, this persistent misalignment compounds and ultimately collapses the generated output. Conversely, when the signals are aligned, they persistently steer the generation toward a semantically consistent solution. This explains why a spatial-temporal attack, which perturbs only the initial latent state but leaves the semantic conditioning intact, is easily overridden by an aligned prompt.

\begin{figure*}[t!]
  \centering
  \includegraphics[width=\linewidth]{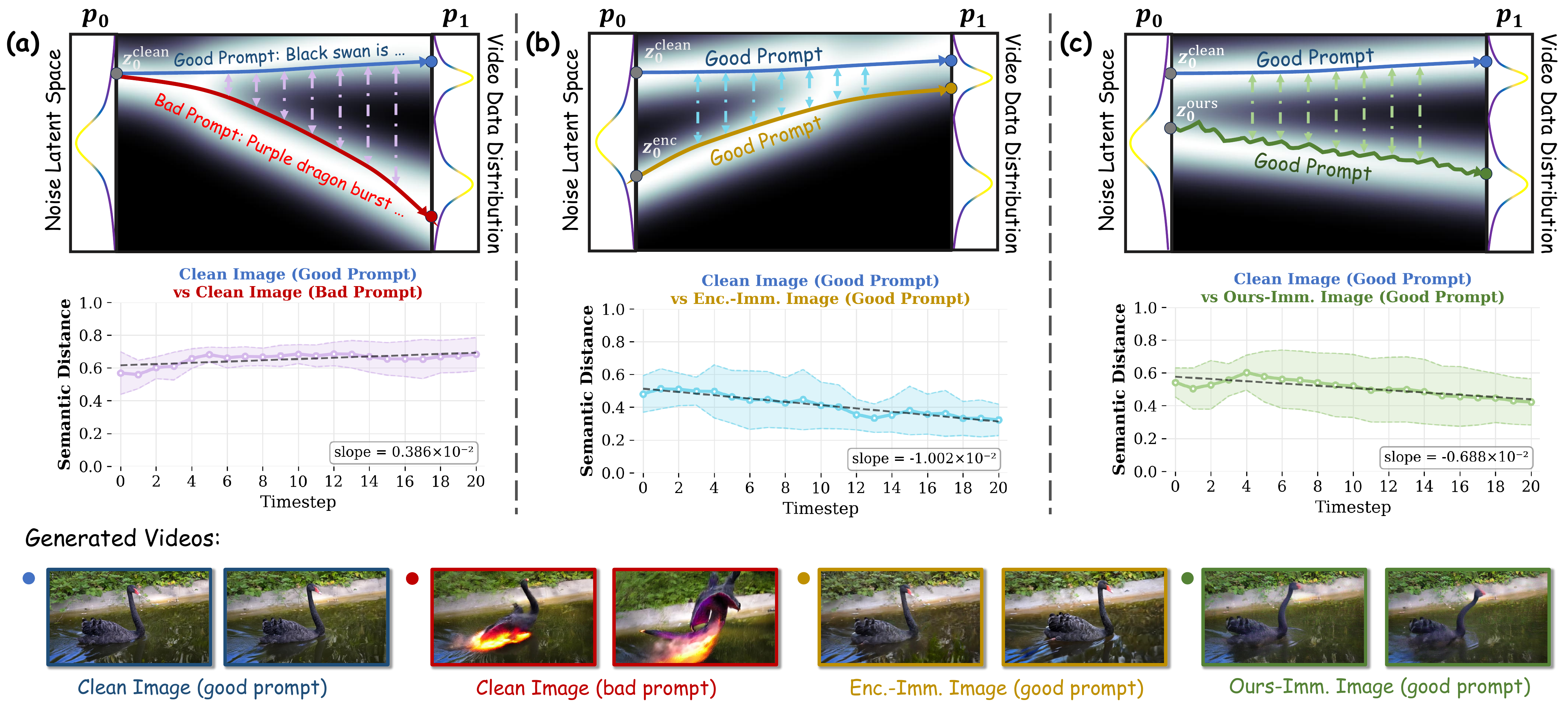}
  \caption{\textit{\textbf{Semantic Conditioning Override.}} We analyze generative trajectories in the noise latent space by measuring semantic distance to a clean baseline (\textcolor[HTML]{4472c4}{Clean Image (Good Prompt)}) at each timestep. (a) A mismatched prompt (\textcolor[HTML]{c00000}{Clean Image (Bad Prompt)}) naturally diverges from the baseline (positive slope). (b) An encoder-only spatial attack (\textcolor[HTML]{bf9000}{Enc.-Imm. Image (Good Prompt)}) is overridden by the continuous semantic guidance of a good prompt, rapidly pulling the trajectory back toward the clean state (steep negative slope). (c) Our joint approach (\textcolor[HTML]{548235}{Ours-Imm. Image (Good Prompt)}) actively counteracts the semantic guidance, maintaining a flatter slope and sustaining the disruption across the generative process.}
  \label{fig:semantic_override}
\end{figure*}

To understand this conditioning dominance empirically, we analyze the generative trajectories in the noise latent space under different text prompts and input images, as illustrated in Fig.~\ref{fig:semantic_override}. Specifically, at each timestep, we track two parallel states: a reference latent (a clean image paired with a good prompt, \ie, one that is aligned with the image, colored blue) and a compared latent (colored red, yellow, or green). For each state and at each timestep, we compute the estimated ground-truth latent and decode it into pixel space. We then quantify trajectory divergence by calculating the Semantic Distance, defined as the cosine distance between the DINO~\cite{caron2021emerging} feature embeddings of these two decoded images.

First, we observe the inherent steering power of the text conditioning. As shown in Fig.~\ref{fig:semantic_override}(a), pairing a clean image with a bad prompt (swan image paired with dragon prompt) causes the generative trajectory to steadily diverge from the baseline, exhibiting a positive slope ($0.386 \times 10^{-2}$) in the change of Semantic Distance from the reference trajectory. The resulting video (Fig. \ref{fig:semantic_override}, red icon) does not clearly resemble either the swan or the dragon. Second, we examine the failure of spatial-only attacks. If an adversarial perturbation is applied exclusively to the encoder, the initial latent state ($z_0^{\mathrm{enc}}$) is successfully corrupted. However, as seen in Fig.~\ref{fig:semantic_override}(b), the continuous injection of a good prompt acts as a strong corrective force during iterative denoising, leading to a steep negative slope ($-1.002 \times 10^{-2}$) of the Semantic Distance. This semantic guidance effectively washes out the initial spatial attack, pulling the corrupted trajectory back toward the clean state.

This constitutes a second structural bottleneck: continuous semantic conditioning override is so dominant that spatial-only perturbations cannot reliably prevent unauthorized video synthesis. To sustain disruption, a defense must explicitly attack the semantic conditioning pathway. As Fig.~\ref{fig:semantic_override}(c) demonstrates, our joint immunization is optimized to actively counteract the good prompt's guidance. This approach significantly dampens the corrective pull, maintaining a much flatter slope ($-0.688 \times 10^{-2}$) and successfully diverting the generated output from the clean baseline.

\subsection{Immune2V Framework: Dual-Stream Joint Optimization}
\label{sec:solution}

In light of these observations, we develop \textbf{Immune2V}, an immunization framework designed to disrupt dual-stream I2V models by by jointly optimizing two complementary attacks: i) a \textbf{Spatial-Temporal Attack} that enforces slice-wise latent divergence across the entire temporal axis of the encoded pseudo-video via a temporally-balanced VAE-level loss, and ii) a \textbf{Semantic Attack} that redirects the condition-driven denoising trajectory toward collapse-inducing dynamics by manipulating intermediate DiT representations.

\begin{figure}[t]
  \centering
  \includegraphics[width=\linewidth]{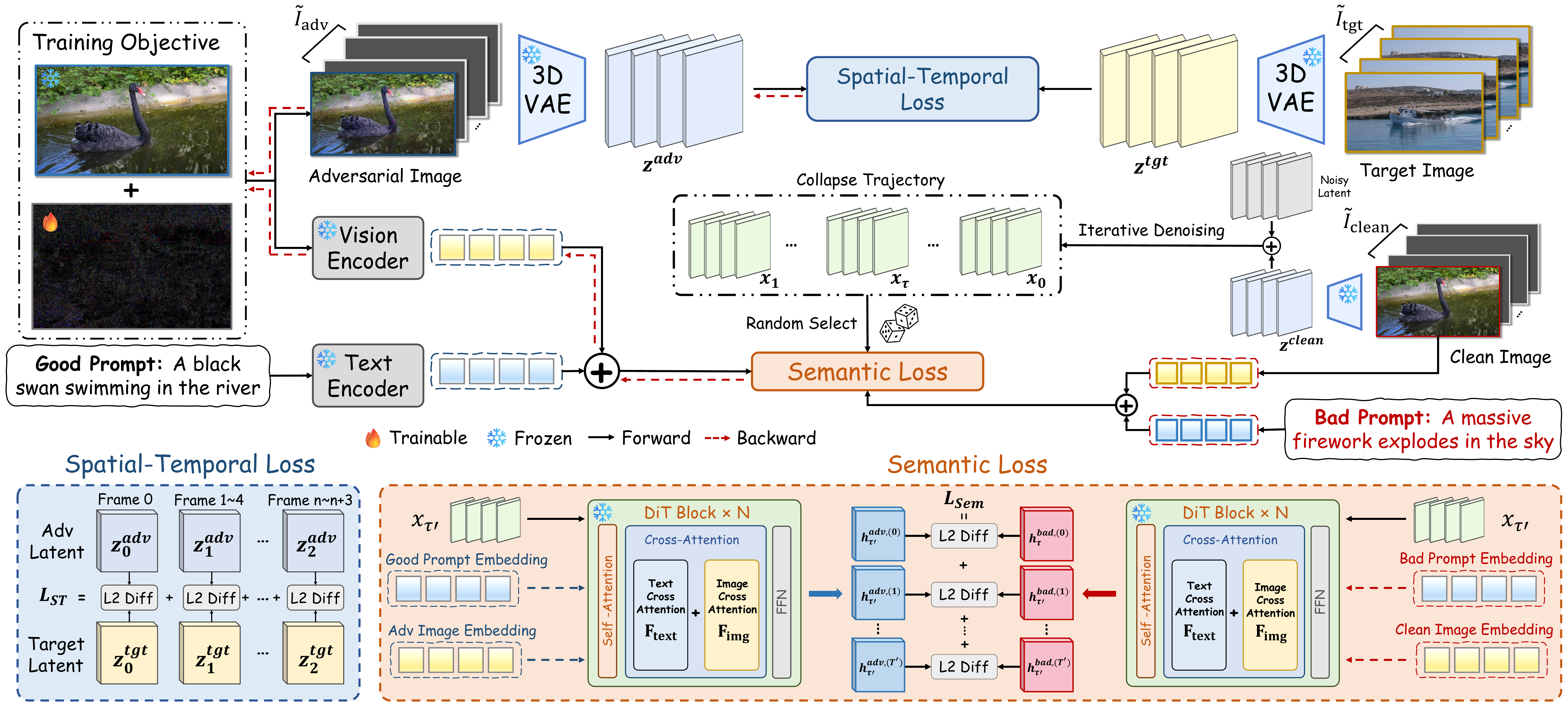}
    \caption{\textit{\textbf{Immune2V Framework.}} Our method simultaneously targets the \textcolor[HTML]{9ECAE1}{spatial-temporal} and \textcolor[HTML]{F4A261}{semantic} streams to ensure persistent disruption. The \textbf{Spatial-Temporal Attack} employs a balanced encoder loss and dense targets to recover vanishing optimization signals across temporal segments. The \textbf{Semantic Attack} hijacks DiT guidance by forcing intermediate representations to mimic a precomputed collapse trajectory, neutralizing the model's iterative semantic correction. Together, these joint perturbations induce severe structural breakdown across the entire generated video.}
  \label{fig:methodology}
\end{figure}

\subsubsection{Spatial-Temporal Attack}
To overcome forward signal decay and backward gradient attenuation, we introduce a temporally-balanced loss imposing active supervision across the entire 3D VAE temporal axis. The key insight is that zero-padded pseudo-video $\tilde{I} = [I, 0, \ldots, 0]$ causes later latent slices to collapse into low-energy states, starving them of useful gradient signals. We address this by replacing the zero-padded target with a \textit{fully-populated} sequence. Specifically, we replicate a reference image $I_{\mathrm{tgt}}$ (\eg, a distinct natural or targeted noisy image) across all temporal positions to form $\tilde{I}_{\mathrm{tgt}} = [I_{\mathrm{tgt}}, I_{\mathrm{tgt}}, \ldots, I_{\mathrm{tgt}}]$. Encoding this sequence through the 3D VAE yields a dense target latent tensor $z^{\mathrm{tgt}} = \mathcal{E}_{\text{enc}}(\tilde{I}_{\mathrm{tgt}})$, guaranteeing every temporal slice contains structured visual information rather than near-zero activations (Fig.~\ref{fig:methodology}, blue dashed box).
We define a slice-wise alignment loss that explicitly pulls each adversarial temporal slice $z_t^{\mathrm{adv}}$ toward its corresponding target slice $z_t^{\mathrm{tgt}}$, computed as the sum of $L_2$ distances over all $T'$ slices:
\begin{equation}
 \mathcal{L}_{\mathrm{ST}} = \sum_{t=1}^{T'} \| z_t^{\mathrm{adv}} - z_t^{\mathrm{tgt}} \|_2.
\end{equation}
By aggregating this loss slice-by-slice, the backward pass accumulates gradient contributions from the entire temporal sequence, circumventing the attenuation bottleneck and forcing the generated video to remain structurally corrupted from the first to the last frame.

\subsubsection{Semantic Attack} As described in Sec.~\ref{sec:fail}, generation collapses when an image is paired with a bad prompt: a clean image combined with an intentionally irrelevant ``bad'' prompt can drive the denoising trajectory into unstable or incoherent states. Our goal is to reproduce this collapse effect while keeping the prompt semantically normal. To achieve this, we design a semantic attack that forces the adversarial image to produce visual embeddings which actively neutralize the coherent text guidance, redirecting the denoising trajectory toward structural collapse.

We first precompute a \textit{collapse trajectory} by running iterative denoising with the clean image $I_{\text{src}}$ and a bad prompt $c_{\text{bad}}$ (\eg, ``A massive firework explodes in the sky''), caching the intermediate noisy latent states $\{ x_\tau \}_{\tau=0}^{1}$ across denoising timesteps.
During adversarial optimization, we fix the text condition to a good prompt $c_{\text{good}}$ (\eg, ``A black swan swimming in the river''), encoding it alongside the adversarial image to obtain $e_{\text{img}}^{\mathrm{adv}} = \mathcal{E}_{\text{img}}(I_{\mathrm{adv}})$ and $e_{\text{txt}}^{\text{good}} = \mathcal{E}_{\text{txt}}(c_{\text{good}})$. At each optimization step, we randomly sample a latent state $x_{\tau'}$ from the cached collapse trajectory and pass it through the first $K$ DiT blocks $f_\theta^{(K)}$ to extract intermediate representations under two parallel conditions: the adversarial setting $h^{\mathrm{adv}}_{\tau'} = f_\theta^{(K)}(x_{\tau'},\, e_{\text{img}}^{\mathrm{adv}},\, e_{\text{txt}}^{\text{good}})$, and the collapse setting $h^{\mathrm{bad}}_{\tau'} = f_\theta^{(K)}(x_{\tau'},\, e_{\text{img}}^{\text{src}},\, e_{\text{txt}}^{\text{bad}})$. By forcing these two representations to match, we ensure that the adversarial image's semantic contribution mimics the effect of the bad prompt, effectively overriding the corrective guidance of the good prompt. We define this alignment loss as the $L_2$ norm of the frame-wise difference over all $T'$ temporal slices:
\begin{equation}
\mathcal{L}_{\mathrm{Sem}}
=
\mathbb{E}_{\tau' \sim \mathcal{U}(0,1)} \left[
\sum_{t=1}^{T'}
\left\|
h^{\mathrm{bad},(t)}_{\tau'}
-
h^{\mathrm{adv},(t)}_{\tau'}
\right\|_2
\right]
\end{equation}
By minimizing $\mathcal{L}_{\mathrm{Sem}}$, the adversarial image embedding learns to cancel out the good prompt's corrective signal, deceiving the iterative generation process and inducing persistent, video-length degradation (see Fig.~\ref{fig:methodology}, orange dashed box).

\subsubsection{Joint Optimization with Dual-Stream Objectives}
We combine the Spatial-Temporal and Semantic Attacks into a unified objective: $\mathcal{L}_{\mathrm{total}} = \lambda_{\mathrm{ST}} \mathcal{L}_{\mathrm{ST}} + \lambda_{\mathrm{Sem}} \mathcal{L}_{\mathrm{Sem}}$, where $\lambda_{\mathrm{ST}}$ and $\lambda_{\mathrm{Sem}}$ control the relative strength of disruption in each stream. To ensure the immunized image remains visually identical to the original, we constrain $\|\delta\|_\infty \le \epsilon$ and optimize via Projected Gradient Descent (PGD)~\cite{madry2017towards}, updating at each iteration as $\delta \leftarrow \Pi_{\|\delta\|_\infty \le \epsilon} (\delta - \alpha \nabla_\delta \mathcal{L}_{\mathrm{total}})$. By jointly targeting both streams, Immune2V simultaneously corrupts the spatial-temporal encoding and hijacks the semantic guidance, inducing severe and persistent degradation across the entire generated video.

\section{Experiments}
\label{sec:experiment}

\begin{table*}[t!]
\centering
\caption{\textbf{\textit{Quantitative comparison.}} All methods are evaluated under the same perturbation budget. \textbf{Bold} and \underline{underline} denote the best and second-best results. PIC: Perceptual Image Consistency, TA: Text-Video Alignment, SC: Subject Consistency, MS: Motion Smoothness, AS: Aesthetic Score, FF: First Frame fidelity.}

\begin{minipage}[c]{0.53\textwidth}
\label{tab:quantitative}
\centering
\subfloat[Automated metrics. Lower is better ($\downarrow$).\label{tab:auto_metrics}]{
\setlength{\tabcolsep}{1.2mm}
\resizebox{\textwidth}{!}{
\begin{tabular}{l|cc|ccc}
\toprule
\multicolumn{1}{c|}{\multirow{3}{*}{\textbf{Methods}}} 
& \multicolumn{2}{c|}{\begin{tabular}{@{}c@{}}Cond.\\[-1pt] Preservation ($\downarrow$)\end{tabular}}
& \multicolumn{3}{c}{\begin{tabular}{@{}c@{}}Video Structural\\[-1pt] Quality ($\downarrow$)\end{tabular}} \\
\cmidrule(lr){2-3} \cmidrule(lr){4-6}
& PIC & TA & SC & MS & AS \\
\midrule
Clean                & 84.97 & 30.55 & 95.88 & 96.26 & 5.24 \\
Random Noise         & 81.15 & 30.28 & 95.49 & 96.58 & 5.27 \\
PhotoGuard-E         & \textbf{73.36} & \underline{29.92} & \underline{93.63} & \underline{95.44} & \underline{4.78} \\
PhotoGuard-D         & 84.23 & 30.25 & 95.12 & 96.50 & 5.07 \\
\midrule
\rowcolor{gray!15}
\textbf{Ours}        & \underline{73.58} & \textbf{29.84} & \textbf{91.71} & \textbf{93.42} & \textbf{4.35} \\
\bottomrule
\end{tabular}
}}
\end{minipage}%
\hfill
\begin{minipage}[c]{0.44\textwidth}
\centering
\subfloat[VLM-as-Judge pairwise win rates (\%). \colorbox{green!15}{Green} backgrounds ({$>$50\%}) and \colorbox{red!15}{red} backgrounds ({$\leq$50\%}) indicate favorable and unfavorable comparisons.\label{tab:vlm_judge}]{
\setlength{\tabcolsep}{1.2mm}
\resizebox{\textwidth}{!}{
\begin{tabular}{l|cccc}
\toprule
\textbf{Ours vs} & TA & SC & MS & FF \\
\midrule
Clean            & \cellcolor{green!15}86.7 & \cellcolor{green!15}86.7 & \cellcolor{green!15}80.0 & \cellcolor{red!15}0.0 \\
Random Noise     & \cellcolor{green!15}60.0 & \cellcolor{green!15}60.0 & \cellcolor{green!15}90.0 & \cellcolor{red!15}26.7 \\
PhotoGuard-E     & \cellcolor{green!15}53.3 & \cellcolor{green!15}56.7 & \cellcolor{green!15}70.0 & \cellcolor{green!15}73.3 \\
PhotoGuard-D     & \cellcolor{green!15}63.3 & \cellcolor{green!15}70.0 & \cellcolor{green!15}60.0 & \cellcolor{red!15}13.3 \\
\bottomrule
\end{tabular}
}}
\end{minipage}
\end{table*}

\subsection{Evaluation Setup}

\subsubsection{Dataset}
Lacking a public I2V adversarial benchmark, we construct an evaluation set from the object-centric DAVIS dataset~\cite{pont20172017}. 
We extract the first frame of 50 videos.
Using GPT-4o-mini, we generate a prompt pair for each image: a \textit{good prompt} describing plausible natural motion, and a \textit{bad prompt} introducing a semantically unrelated scene with conflicting dynamics to act as the collapse-inducing condition. Further details are in the Supplementary Material.

\subsubsection{Baselines}
Since no prior work directly addresses I2V image immunization (the closest method, I2VGuard~\cite{gui2025i2vguard}, is closed-source), we evaluate against three baselines. \textbf{Clean Input} uses the unperturbed image as the quality upper bound. \textbf{Random Noise} applies uniform perturbations bounded by the same $\ell_\infty$ budget $\epsilon$, establishing a lower bound. Finally, we adapt \textbf{PhotoGuard}~\cite{salman2023raising}---an image-level defense minimizing the $\ell_2$ distance to a target---into two I2V variants. \textbf{PhotoGuard-E} targets only the 3D VAE encoder by minimizing the distance between clean and target latents, while \textbf{PhotoGuard-D} backpropagates through the entire diffusion sampling process. For all baselines, we use the frozen Wan 2.1 (I2V-14B-480P)~\cite{wan2025wanopenadvancedlargescale} pipeline. Additional details are in the Supplementary Material.

\begin{figure*}[t!]
\centering
\includegraphics[width=\linewidth]{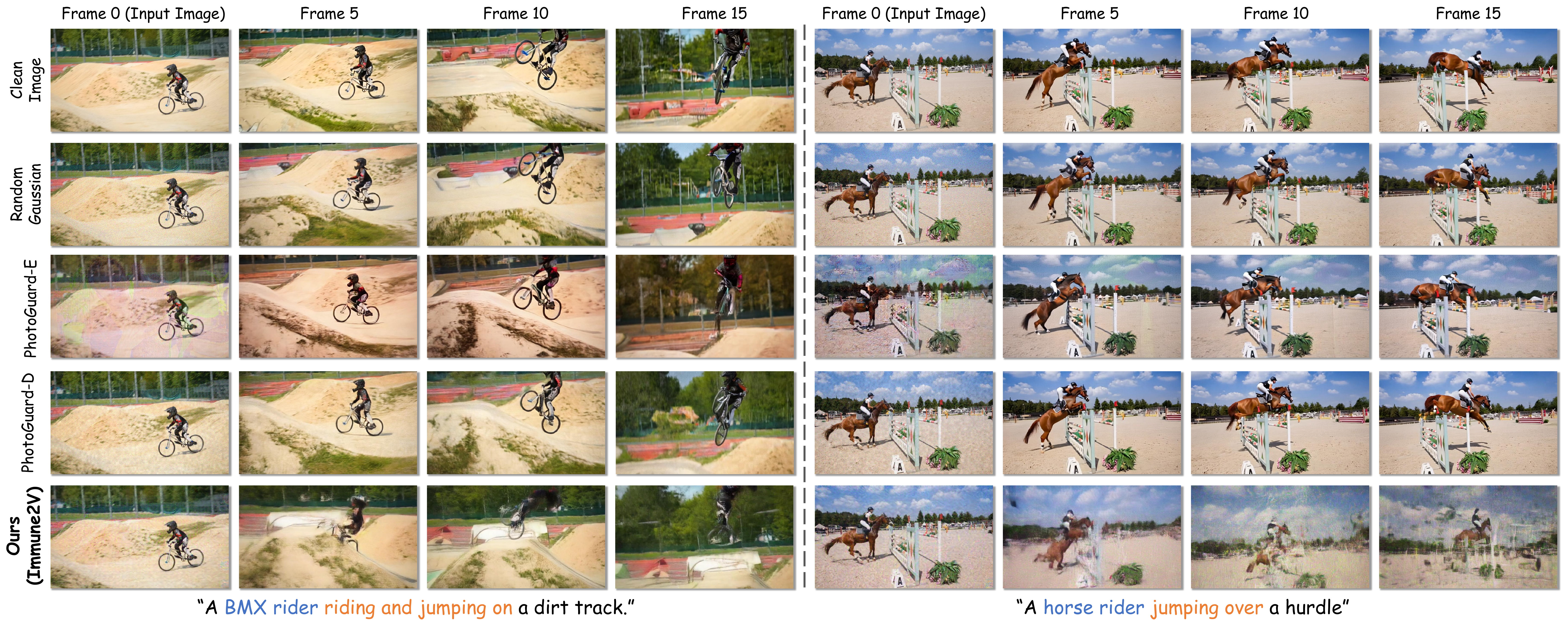}
\caption{\textbf{\textit{Qualitative Comparisons.}} A qualitative comparison across baselines, showing the input frame and subsequent video frames. See the Supplementary  for full videos.}
\label{fig:qualitative}
\end{figure*}

\subsubsection{Automated Quantitative Metrics}
Successful immunization degrades conditioning fidelity and structural quality, evaluated across two categories:
i) \textbf{Condition Preservation} assesses fidelity to conditioning signals. \textit{Perceptual Image Consistency (PIC, $\downarrow$)}~\cite{xing2024dynamicrafter} averages DreamSim~\cite{fu2023dreamsim} similarity between input and generated frames to capture source divergence. \textit{Text Alignment (TA, $\downarrow$)} computes the average cosine similarity between CLIP~\cite{radford2021learning} embeddings of the prompt and each frame for semantic alignment.
ii) \textbf{Video Structural Quality} evaluates temporal coherence and visual plausibility. \textit{Subject Consistency (SC, $\downarrow$)} computes pairwise DINO~\cite{caron2021emerging} similarities across consecutive frames to measure appearance stability. \textit{Motion Smoothness (MS, $\downarrow$)} proxies motion regularity via the reconstruction error of a VFI-based interpolation model predicting intermediate frames. \textit{Aesthetic Score (AS, $\downarrow$)} averages per-frame LAION~\cite{schuhmann2022laion} scores to estimate overall visual quality.

\subsubsection{VLM-as-Judge}
\label{sec:vlm_judge}
To complement the automated metrics, we employ Gemini~3.1 Pro~\cite{google_ai_dev} as a pairwise judge. For each scene, the VLM is presented with our generated video alongside a baseline generated video and asked to determine \emph{which is worse} across the perceptual equivalents of our automated metrics: \textbf{Text Alignment (TA)}, \textbf{Subject Consistency (SC)}, and \textbf{Motion Smoothness (MS)}. Additionally, we evaluate \textbf{First Frame (FF)} distortion, where the VLM determines which initial frame appears more corrupted based on a single-image comparison. We report the percentage of scenes where our output is judged worse for TA, SC, MS, and FF. Further details are in the Supplementary Material.

\begin{figure*}[t!]
    \centering
    \includegraphics[width=\textwidth]{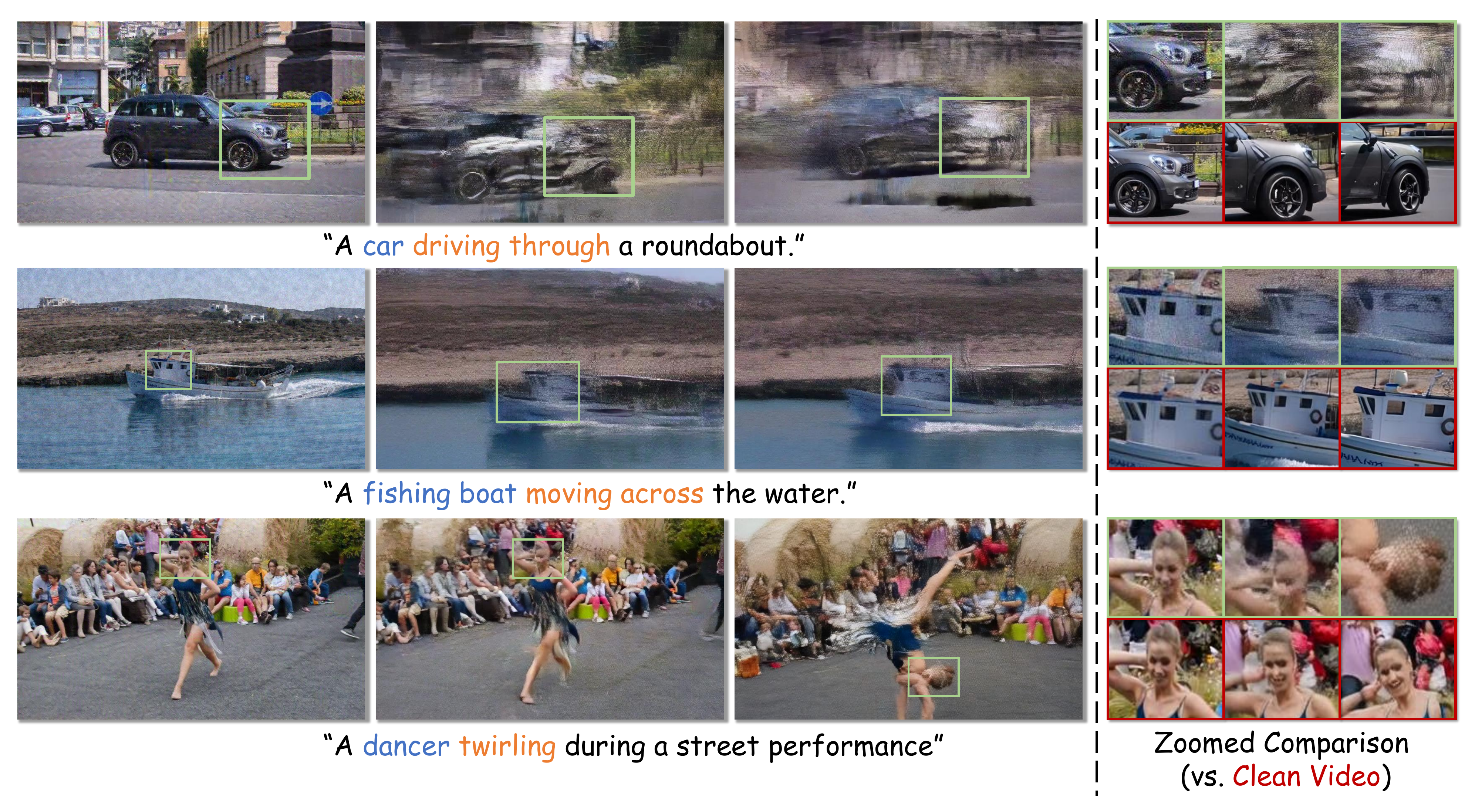}
    \caption{\textbf{\textit{Additional Qualitative Results.}} Sampled frames from the generated videos are shown on the left, with zoomed comparisons against the corresponding clean video frames on the right. Full videos are available in Supplementary.}
    \label{fig:more qualitative}
\end{figure*}

\subsection{Analysis and Discussion}

\subsubsection{Automated Quantitative Evaluation}

Table~\ref{tab:auto_metrics} compares our method against existing image-level defenses. Our approach yields the strongest degradation in both conditioning preservation and video structural quality, achieving the lowest scores for text alignment, subject consistency, motion smoothness, and aesthetics. PhotoGuard-E slightly outperforms us on PIC because its objective directly targets the VAE, introducing strong distortions at the first frame. However, these perturbations remain largely confined to the encoding stage and fail to propagate effectively through the denoising generation process, limiting their impact on the final video.

\subsubsection{VLM-as-Judge}
Table~\ref{tab:vlm_judge}'s VLM evaluation corroborates the automated metrics. Regarding \textit{First Frame (FF)} fidelity, PhotoGuard-E introduces severe visual distortions because its fast-converging, encoder-only loss overfits to the input. Conversely, our method maintains low FF distortion, keeping the initial adversarial image perceptually intact. For \textit{Text Alignment (TA)}, our semantic attack successfully neutralizes coherent text guidance, forcing the content to diverge from the prompt. In terms of \textit{Subject Consistency (SC)}, our temporally-balanced attack reliably breaks the target's appearance over time, preventing the model from recovering the subject in later frames. Finally, the VLM consistently identifies our approach as the most successful attack across all these metrics, including \textit{Motion Smoothness (MS)}, confirming that our joint dual-stream optimization effectively induces a structural collapse throughout the video.

\begin{table*}[t]
\centering
\caption{\textbf{\textit{Ablation Study.}} We ablate individual loss components and noise budgets from our full model. $\mathcal{L}_{\text{ST}}$: spatial-temporal loss, $\mathcal{L}_{\text{Sem}}$: semantic loss. All metrics are $\downarrow$ (lower is better). Our full model is highlighted in gray and repeated per section. Deltas are relative to our full model: \textcolor{winGreen}{green} = improved, \textcolor{lossRed}{red} = degraded.}
\label{tab:ablation}
\vspace{-3mm}
\setlength{\tabcolsep}{6pt}
\renewcommand{\arraystretch}{1.15}
\resizebox{\textwidth}{!}{
\begin{tabular}{@{} ccc cc ccc @{}}
\toprule
\multicolumn{3}{@{}c}{\textbf{Configuration}}
& \multicolumn{2}{c}{\textbf{Condition Preservation} $\downarrow$}
& \multicolumn{3}{c@{}}{\textbf{Video Structural Quality} $\downarrow$} \\
\cmidrule(lr){1-3}
\cmidrule(lr){4-5}
\cmidrule(lr){6-8}
$\mathcal{L}_{\text{ST}}$ & $\mathcal{L}_{\text{Sem}}$ & $\epsilon$
& PIC (\%) 
& Text Align.
& Subject Cons.\ (\%)
& Motion Smooth.\ (\%)
& AS \\
\midrule
\multicolumn{8}{c}{\cellcolor{RoyalBlue!8}\textbf{Ablation on Loss Components} (fixed $\epsilon = 32/255$)} \\
\addlinespace[2pt]

\textcolor{winGreen}{\ding{51}} & \textcolor{lossRed}{\ding{55}} & 32/255 
& 75.86 {\scriptsize \textcolor{lossRed}{(+2.28)}} 
& \textcolor{winGreen}{29.74} {\scriptsize \textcolor{lossRed}{(-0.10)}} 
& 92.12 {\scriptsize \textcolor{lossRed}{(+0.41)}} 
& 94.04 {\scriptsize \textcolor{lossRed}{(+0.62)}} 
& 4.87 {\scriptsize \textcolor{lossRed}{(+0.51)}} \\

\rowcolor{gray!6}
\textcolor{lossRed}{\ding{55}} & \textcolor{winGreen}{\ding{51}} & 32/255 
& 82.34 {\scriptsize \textcolor{lossRed}{(+8.76)}} 
& 29.95 {\scriptsize \textcolor{lossRed}{(+0.11)}} 
& 93.68 {\scriptsize \textcolor{lossRed}{(+1.97)}} 
& 94.27 {\scriptsize \textcolor{lossRed}{(+0.85)}} 
& 5.02 {\scriptsize \textcolor{lossRed}{(+0.66)}} \\

\rowcolor{gray!15}
\textcolor{winGreen}{\ding{51}} & \textcolor{winGreen}{\ding{51}} & \textbf{32/255} 
& \textcolor{winGreen}{\textbf{73.58}} 
& 29.84 
& \textcolor{winGreen}{\textbf{91.71}} 
& \textcolor{winGreen}{\textbf{93.42}} 
& \textcolor{winGreen}{\textbf{4.36}} \\

\multicolumn{8}{c}{\cellcolor{Plum!8}\textbf{Ablation on Perturbation Budget} (fixed $\mathcal{L}_{\text{ST}} + \mathcal{L}_{\text{Sem}}$)} \\
\addlinespace[2pt]

\textcolor{winGreen}{\ding{51}} & \textcolor{winGreen}{\ding{51}} & 8/255 
& 84.41 {\scriptsize \textcolor{lossRed}{(+10.83)}} 
& 30.81 {\scriptsize \textcolor{lossRed}{(+0.97)}} 
& 93.19 {\scriptsize \textcolor{lossRed}{(+1.48)}} 
& 95.72 {\scriptsize \textcolor{lossRed}{(+2.30)}} 
& 5.14 {\scriptsize \textcolor{lossRed}{(+0.78)}} \\

\rowcolor{gray!6}
\textcolor{winGreen}{\ding{51}} & \textcolor{winGreen}{\ding{51}} & 16/255 
& 83.63 {\scriptsize \textcolor{lossRed}{(+10.05)}} 
& 29.96 {\scriptsize \textcolor{lossRed}{(+0.12)}} 
& 92.62 {\scriptsize \textcolor{lossRed}{(+0.91)}} 
& 94.83 {\scriptsize \textcolor{lossRed}{(+1.41)}} 
& 5.03 {\scriptsize \textcolor{lossRed}{(+0.67)}} \\

\rowcolor{gray!15}
\textcolor{winGreen}{\ding{51}} & \textcolor{winGreen}{\ding{51}} & \textbf{32/255} 
& \textcolor{winGreen}{\textbf{73.58}} 
& 29.84 
& 91.71 
& 93.42
& 4.36 \\

\textcolor{winGreen}{\ding{51}} & \textcolor{winGreen}{\ding{51}} & 64/255 
& 77.50 {\scriptsize \textcolor{lossRed}{(+3.92)}} 
& \textcolor{winGreen}{27.77} {\scriptsize \textcolor{winGreen}{(-2.07)}} 
& \textcolor{winGreen}{89.14} {\scriptsize \textcolor{winGreen}{(-2.57)}} 
& \textcolor{winGreen}{92.11} {\scriptsize \textcolor{winGreen}{(-1.31)}} 
& \textcolor{winGreen}{4.13} {\scriptsize \textcolor{winGreen}{(-0.23)}} \\
\bottomrule
\end{tabular}
}
\end{table*}

\subsubsection{Qualitative Comparisons}

Fig.~\ref{fig:qualitative} compares our method against baselines. While clean inputs and baselines yield coherent videos with plausible motion, our approach induces progressive structural collapse (\eg, the BMX rider's geometry distorts mid-sequence, and the horse rider loses anatomical coherence) that persists across frames. Notably, PhotoGuard-E leaves visible perturbation traces on the input image due to its fast-converging encoder-only loss, yet this corruption fails to transfer to the generated video. Our joint dual-stream approach avoids these failure modes by simultaneously targeting spatial-temporal encoding and semantic conditioning.
Fig.~\ref{fig:more qualitative} displays additional attacked examples. Across diverse scenes (humans and objects), the generated videos exhibit progressive structural breakdown, deteriorating object geometry and fine details into unstable textures. Zoomed comparisons with clean videos highlight these severe distortions within the bounding-boxed regions.

\subsubsection{Ablation Study}

We ablate the loss design and perturbation budget (Table~\ref{tab:ablation}). Applying only the semantic loss yields the weakest degradation, as its perturbed conditional guidance diminishes along the denoising trajectory due to temporal attenuation (Sec.~\ref{sec:fail}). The spatial-temporal loss alone yields stronger degradation, while combining both produces the most consistent structural collapse, proving joint optimization is most effective. For the perturbation budget, small magnitudes (\eg, $8/255$, $16/255$) provide insufficient attack signals,whereas larger values (\eg, $64/255$) produce stronger distortion in the videos while also introducing noticeable artifacts in the input image. Thus, to balance attack effectiveness and perceptual fidelity of the input image, we adopt $\epsilon = 32/255$ as a practical setting.

\section{Conclusion}
\label{sec:conclusion}

We present \textbf{Immune2V}, a dual-stream adversarial immunization framework that protects images from unauthorized Image-to-Video (I2V) generation. We identify two structural bottlenecks in modern dual-stream I2V models—temporal perturbation attenuation in the spatial-temporal encoder and conditioning override in the semantic pathway—that fundamentally limit naively extended image-level defenses. To address these challenges, Immune2V introduces a Spatial-Temporal Attack enforcing temporally-balanced latent supervision in the 3D VAE, and a Semantic Attack redirecting cross-attention trajectories toward collapse-inducing dynamics. Jointly optimizing these objectives under strict imperceptibility constraints produces a single input perturbation that induces persistent structural and motion degradation across the entire generated video. Extensive evaluations demonstrate that our method effectively disrupts state-of-the-art dual-stream I2V models while preserving the protected image's visual fidelity, highlighting the necessity of video-aware protection mechanisms in the era of generative synthesis.

\newpage
\bibliographystyle{unsrt}
\bibliography{main}

\newpage






\appendix

\begingroup
\hypersetup{
    citecolor=blue,
    linkcolor=blue
}

\section*{Appendix Contents}

\startcontents[appendices]
\printcontents[appendices]{l}{1}{\setcounter{tocdepth}{2}}
\endgroup




\newpage
\section{Immune2V Algorithm Details}

\subsection{Method Overview (Recap)}

\textbf{Immune2V} is designed for image immunization in Image-to-Video (I2V) generation, where the input image influences video generation process through two conditioning streams: a spatial-temporal stream and a semantic stream. By disrupting both pathways, the generation dynamics can be effectively destabilized. The goal is to optimize a small perturbation on the input image that simultaneously attacks both streams, causing structural collapse in the spatial-temporal latent trajectory and semantic interference during denoising.

Algorithm~\ref{alg:immune2v} summarizes the overall optimization procedure.

\begin{algorithm}[ht]
\caption{Immune2V: Image Immunization for Dual-Stream I2V Models}
\label{alg:immune2v}
\textbf{Input}: Clean image $I$, target image $I_{\text{tgt}}$, good prompt $c_{\text{good}}$, bad prompt $c_{\text{bad}}$, 
3D VAE encoder $\mathcal{E}_{\text{enc}}$, image encoder $\mathcal{E}_{\text{img}}$, text encoder $\mathcal{E}_{\text{txt}}$, 
DiT backbone $f_\theta$,  
step size $\alpha$, perturbation budget $\epsilon$, loss weights $\lambda_{\mathrm{ST}}, \lambda_{\mathrm{Sem}}$, optimization steps $N$
\begin{algorithmic}[1]

\STATE \textbf{Initialize} $\delta \sim \mathcal{U}(-\epsilon, \epsilon)$
\STATE $\tilde{I}_{\text{tgt}} \gets [I_{\text{tgt}}, I_{\text{tgt}}, \ldots, I_{\text{tgt}}]$
\STATE $z^{\text{tgt}} \gets \mathcal{E}_{\text{enc}}(\tilde{I}_{\text{tgt}})$ \hfill $\triangleright$ Dense temporal target
\STATE Run denoising with $(I, c_{\text{bad}})$ and cache collapse trajectory $\{x_\tau\}_{\tau=0}^{1}$

\FOR{$i = 1$ to $N$}
    \STATE $I_{\text{adv}} \gets I + \delta$
    
    \STATE $\tilde{I}_{\text{adv}} \gets [I_{\text{adv}}, 0, \ldots, 0]$
    \STATE $z^{\text{adv}} \gets \mathcal{E}_{\text{enc}}(\tilde{I}_{\text{adv}})$
    
    \STATE $\mathcal{L}_{\mathrm{ST}} \gets \sum_{t=1}^{T'} \|z_t^{\text{adv}} - z_t^{\text{tgt}}\|_2$ \hfill $\triangleright$ Spatial-temporal loss

    \STATE \textcolor{gray}{\hdashrule{\linewidth}{0.3pt}{3pt 2pt}}
    
    \STATE \textbf{Sample} $\tau' \sim \mathcal{U}(0,1)$
    \STATE $x_{\tau'} \gets$ cached collapse latent
    
    \STATE $e_{\text{img}}^{\text{adv}} \gets \mathcal{E}_{\text{img}}(I_{\text{adv}})$, $e_{\text{img}}^{\text{src}} \gets \mathcal{E}_{\text{img}}(I)$; $e_{\text{txt}}^{\text{good}} \gets \mathcal{E}_{\text{txt}}(c_{\text{good}})$, $e_{\text{txt}}^{\text{bad}} \gets \mathcal{E}_{\text{txt}}(c_{\text{bad}})$
    
    \STATE $h_{\tau'}^{\text{adv}} \gets f_\theta^{(K)}(x_{\tau'}, e_{\text{img}}^{\text{adv}}, e_{\text{txt}}^{\text{good}})$, $h_{\tau'}^{\text{bad}} \gets f_\theta^{(K)}(x_{\tau'}, e_{\text{img}}^{\text{src}}, e_{\text{txt}}^{\text{bad}})$
    
    \STATE $\mathcal{L}_{\mathrm{Sem}} \gets \sum_{t=1}^{T'} \|h_{\tau'}^{\text{bad},(t)} - h_{\tau'}^{\text{adv},(t)}\|_2$ \hfill $\triangleright$ Semantic loss
    
    \STATE $\mathcal{L} \gets \lambda_{\mathrm{ST}}\mathcal{L}_{\mathrm{ST}} + \lambda_{\mathrm{Sem}}\mathcal{L}_{\mathrm{Sem}}$
    
    \STATE $\delta \gets \delta - \alpha \nabla_\delta \mathcal{L}$
    \STATE $\delta \gets \Pi_{\|\delta\|_\infty \le \epsilon}(\delta)$ \hfill $\triangleright$ PGD optimization
\ENDFOR

\STATE \textbf{return} $I_{\text{adv}} = I + \delta$
\end{algorithmic}
\end{algorithm}

\newpage
\subsection{Model-Specific Implementation Notes}

While the formulation of Immune2V is intentionally general, modern I2V systems may instantiate the dual-stream conditioning structure with slightly different architectural components. 
In practice, the spatial-temporal pathway is typically implemented using a 3D VAE that converts video frames into a latent representation used to initialize the diffusion or flow-based denoising process. 
Some implementations compress the temporal dimension during encoding \cite{yang2024cogvideox, wan2025wanopenadvancedlargescale}, while others preserve the full frame sequence without temporal downsampling \cite{xing2024dynamicrafter}. 
Our spatial-temporal objective is compatible with both designs, as it operates directly on the resulting latent representations.
In Wan 2.1, the encoded latent is additionally combined with Gaussian noise through a small MLP to produce the initial latent for the diffusion process. This design mainly serves as a specific initialization strategy and does not change the fundamental role of the spatial-temporal conditioning pathway.

For the semantic conditioning pathway, the input image is usually encoded by a pretrained vision-language encoder. In many recent pipelines this is a CLIP-based image encoder \cite{radford2021learning}, which provides rich semantic features for cross-attention during generation.
The text prompt is commonly processed by a large language encoder such as T5 \cite{raffel2020exploring} or a similar transformer-based text model. 
The resulting image and text embeddings are then fused through cross-attention inside the denoising backbone, which may be implemented using U-Net \cite{ronneberger2015u} or DiT-based \cite{peebles2023scalable} architectures.

Immune2V does not rely on the specific choice of encoders or backbone architecture. 
As a result, the proposed optimization procedure can be applied to a wide range of dual-stream I2V generation frameworks with minimal modification.

\newpage

\section{Additional Experiment Details}
\subsection{Attack Optimization Details \& Hyperparameters}

In the main paper, all experiments are conducted using the Wan~2.1 (I2V-14B-480P) pipeline~\cite{wan2025wanopenadvancedlargescale} with frozen model parameters. The input resolution is set to $480 \times 832$, and the number of generated frames is fixed to $17$. We optimize input-space perturbations using Projected Gradient Descent (PGD) for $500$ iterations under an $\ell_\infty$ constraint of $\epsilon = 32/255$ and a step size $\alpha = 0.005$.

During optimization, a collapse trajectory needs to be generated. We set the number of denoising steps to $25$, which defines the trajectory length. In practice, the preprocessing classifier-free guidance scale (CFG scale) can be adjusted (e.g., $1$, $3$, or $5$) to control the strength of semantic conditioning.

To target the early stage of video layout formation, the attack is restricted to the first $10$ timesteps out of the $25$ total inference steps, corresponding to the region closest to the pure-noise initialization stage. For latent feature extraction, we use the outputs of the first $K=3$ DiT blocks to compute the semantic attack objective. The coefficients $\lambda_{ST}$ and $\lambda_{Sem}$ control the relative strengths of the spatial-temporal and semantic losses. Since the semantic loss typically has a larger magnitude and higher variance during optimization, we set $\lambda_{ST}=1$ and $\lambda_{Sem}=0.125$ to balance the two objectives. These weights can be adjusted to emphasize either structural collapse or semantic interference depending on the desired attack behavior.

During evaluation, the CFG scale is set to one of $\{1,3,5\}$ depending on the generation configuration, although using the same value as in the preprocessing stage is recommended for consistency.

With gradient checkpointing enabled during the backward pass, immunizing a single image requires approximately $25$ minutes on one NVIDIA A100 (80GB) GPU. Video generation during testing takes approximately $1$ minute per sample. With larger GPU memory, the optimization can be extended to longer video with more frames. Exploring more memory-efficient optimization strategies to support such settings is left for future work.

All hyperparameters used in our experiments are summarized in Table~\ref{tab:all_hyperparameters}.

\begin{table}[h!]
\centering
\caption{Hyperparameters for Immune2V optimization and generation configuration.}
\label{tab:all_hyperparameters}
\resizebox{\linewidth}{!}{
\begin{tabular}{|c|c|c|c|c|c|c|c|c||c|c|}
\hline
\multicolumn{9}{|c||}{\textbf{Attack Optimization}} & \multicolumn{2}{c|}{\textbf{Generation}} \\
\hline
\textbf{Resolution} & \textbf{Frames} & \textbf{PGD Iters} & \textbf{Budget $\epsilon$} & \textbf{Step $\alpha$} & \textbf{Traj. Length} & \textbf{DiT ($K$)} & $\bm{\lambda_{ST}}$ & $\bm{\lambda_{Sem}}$ & \textbf{CFG Scale} & \textbf{Denoising Steps} \\
\hline
$480 \times 832$ & 17 & 500 & $32/255$ & 0.005 & 25 & 3 & 1 & 0.125 & $\{1,3,5\}$ & 25 \\
\hline
\end{tabular}
}
\end{table}

\newpage
\subsection{Baselines}

In both quantitative and qualitative experiments, we compare our method with two variants of PhotoGuard~\cite{salman2023raising}: PhotoGuard-E and PhotoGuard-D. In the original work, these correspond to the encoder-level and diffusion-level objectives, respectively, defined as
\begin{align}
\delta_{\text{encoder}}
&=
\arg\min_{\|\delta\|_\infty \le \epsilon}
\left\| \mathcal{E}_\text{enc}(I+\delta) - z^{\text{tgt}} \right\|_2^2
\\
\delta_{\text{diffusion}}
&=
\arg\min_{\|\delta\|_\infty \le \epsilon}
\left\| f(I+\delta) - I_{\text{tgt}} \right\|_2^2
\end{align}

For the PhotoGuard-E experiments, we follow the original encoder-level objective and use the same target image as in our method. To adapt the formulation to the video pipeline, the target image is padded with black frames.

For the PhotoGuard-D setting, directly backpropagating through the full video diffusion process (17 frames) is computationally infeasible in our setting due to memory constraints. Therefore, we adopt a simplified image-level formulation for PhotoGuard-D. Specifically, the number of frames is set to 1 and the number of denoising steps is reduced to 10, while using the same target image as in our method.


\newpage
\section{Dataset and Evaluation}

\subsection{Dataset Construction}

The DAVIS dataset \cite{pont20172017} contains high-quality videos with clearly defined foreground objects. We select the first frame from a subset of DAVIS videos where the main object is clearly visible and belongs to common semantic categories such as humans, animals, or everyday objects. To increase diversity, we additionally include several images collected from other public sources that satisfy the same selection criteria. In total, the dataset contains 50 source images. For each source image, another image from the dataset is randomly selected and used as the target image.

For each image, we generate a corresponding \textit{good prompt} using ChatGPT~4o-mini. The prompt is designed to remain semantically consistent with the main object in the image and describes a natural action that the object could plausibly perform. This ensures that the prompt is well aligned with the visual content while still inducing realistic motion dynamics.

We also generate a set of \textit{bad prompts}. In contrast, these prompts intentionally describe objects or scenarios that are semantically unrelated to the original image content, creating a strong mismatch between the image and the textual conditioning. Such prompts encourage the generation model to produce conflicting outputs.

The prompt templates used to generate both the good and bad prompts are provided in Fig.~\ref{fig:prompt_template}.

\begin{figure}[ht]
\centering

\begin{tcolorbox}[title=Good and Bad Prompt Generation Template]

You are an expert prompt designer for Wan 2.1 image-to-video generation.  
You are given an image containing one main object (e.g., an animal, a person, a vehicle, etc.).

Generate two short I2V prompts.

\textbf{Prompt 1 (Good Prompt)}
\begin{itemize}
\item Use the exact main object from the image.
\item Describe a realistic, physically plausible motion over time.
\item The motion should be natural, smooth, and consistent with the object's identity.
\item The movement should show clear momentum and speed progression (avoid static motion).
\item Focus on temporal progression and describe how the motion evolves step by step.
\end{itemize}

\textbf{Prompt 2 (Bad Prompt)}
\begin{itemize}
\item Use a completely unrelated object.
\item The motion should be drastically different in category and dynamics.
\item The motion should be intense, exaggerated, high-energy, and visually dramatic.
\item The contrast between the two prompts should be very strong.
\end{itemize}

\end{tcolorbox}

\caption{Prompt templates used to generate good and bad prompts.}
\label{fig:prompt_template}

\end{figure}


\subsection{VLM-as-a-Judge}

In addition to the automated metrics reported in the main paper, we perform a complementary perceptual evaluation using a Vision-Language Model (VLM) as a pairwise judge. While quantitative metrics capture numerical degradation, they do not fully reflect high-level semantic violation or physical implausibility introduced by adversarial perturbations. In practice, simply degrading generation quality does not necessarily imply a successful attack, as stochastic sampling or poorly conditioned prompts may also reduce numerical scores. Therefore, we employ Gemini 3.1 Pro as an external evaluator to provide a model-based perceptual comparison on a subset of 30 images of our test-set.

For each scene, the VLM is presented with two generated videos: one produced from the immunized image and one from the baseline or clean image. The VLM is asked to identify which video appears worse under three criteria corresponding to the perceptual counterparts of our automated metrics: Text Alignment (TA), Subject Consistency (SC), and Motion Smoothness (MS). In addition, we evaluate First Frame (FF) distortion through a single-image comparison of the initial frames. The judge outputs a binary preference indicating which sample is worse for each criterion, and we report the percentage of scenes where our output is judged to be worse. Prompt templates used for the VLM evaluation are provided in Fig.~\ref{fig:vlmjudge}.

\begin{figure}[ht]
\centering

\begin{tcolorbox}[title=VLM Judge Evaluation Prompts]

\textbf{Text Alignment}

You are an expert video quality evaluator.  
You will see two short videos: Video A and Video B.

The videos were generated from the same text description.  
Decide which video \textbf{less matches} the text description — i.e., worse alignment with the described actions, subject behaviour, and overall semantics.

Reply with only the single letter \textbf{A} or \textbf{B}.

\vspace{0.5em}

\textbf{Subject Consistency}

You are an expert video quality evaluator.  
You will see two short videos: Video A and Video B.

Decide which video has \textbf{less consistent subject across frames} — e.g., more flickering, morphing, identity drift, or distortion of the main subject.

Reply with only the single letter \textbf{A} or \textbf{B}.

\vspace{0.5em}

\textbf{Motion Preservation}

You are an expert video quality evaluator.  
You will see two short videos: Video A and Video B.

Decide which video has \textbf{less natural motion} — e.g., jitter, unnatural trajectories, sudden jumps, or physically implausible movement.

Reply with only the single letter \textbf{A} or \textbf{B}.

\vspace{0.5em}

\textbf{First Frame Quality}

You are an expert image quality assessor.  
You are shown two images: Image A and Image B. Both are the first frames of videos generated from the same source photograph.

Decide which image is \textbf{more distorted} — e.g., more visual artifacts, noise, identity distortion, or less natural appearance.

Reply with only the single letter \textbf{A} or \textbf{B}.

\end{tcolorbox}

\caption{Prompt templates used by \textbf{VLM Judge} for automated pairwise evaluation of generated videos and first-frame image quality.}
\label{fig:vlmjudge}

\end{figure}

\clearpage
\section{Additional Results on other I2V Models}


\begin{figure}[ht]
    \centering
    \includegraphics[width=0.9\linewidth]{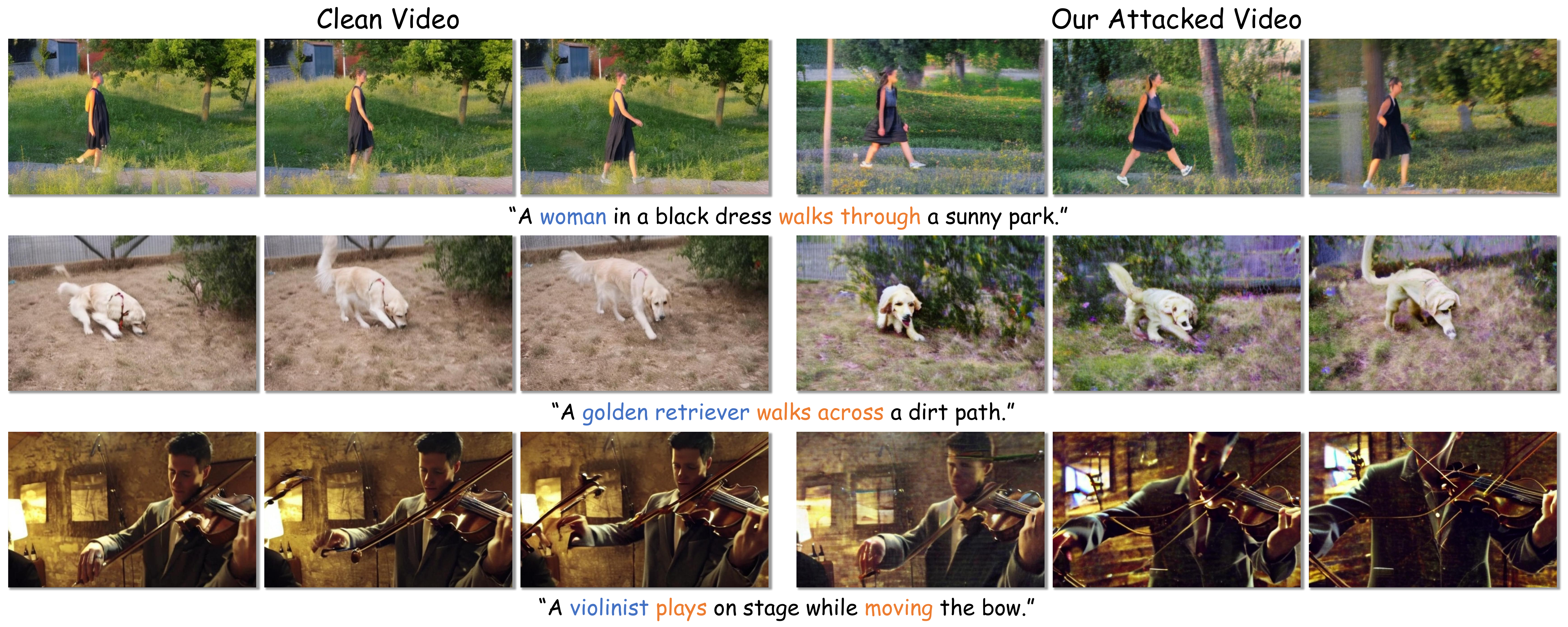}
    \caption{Clean and attacked video results on DynamiCrafter using Immune2V.}
    \label{fig:dynamicrafter}
\end{figure}


\begin{figure}[ht]
    \centering
    \includegraphics[width=0.9\linewidth]{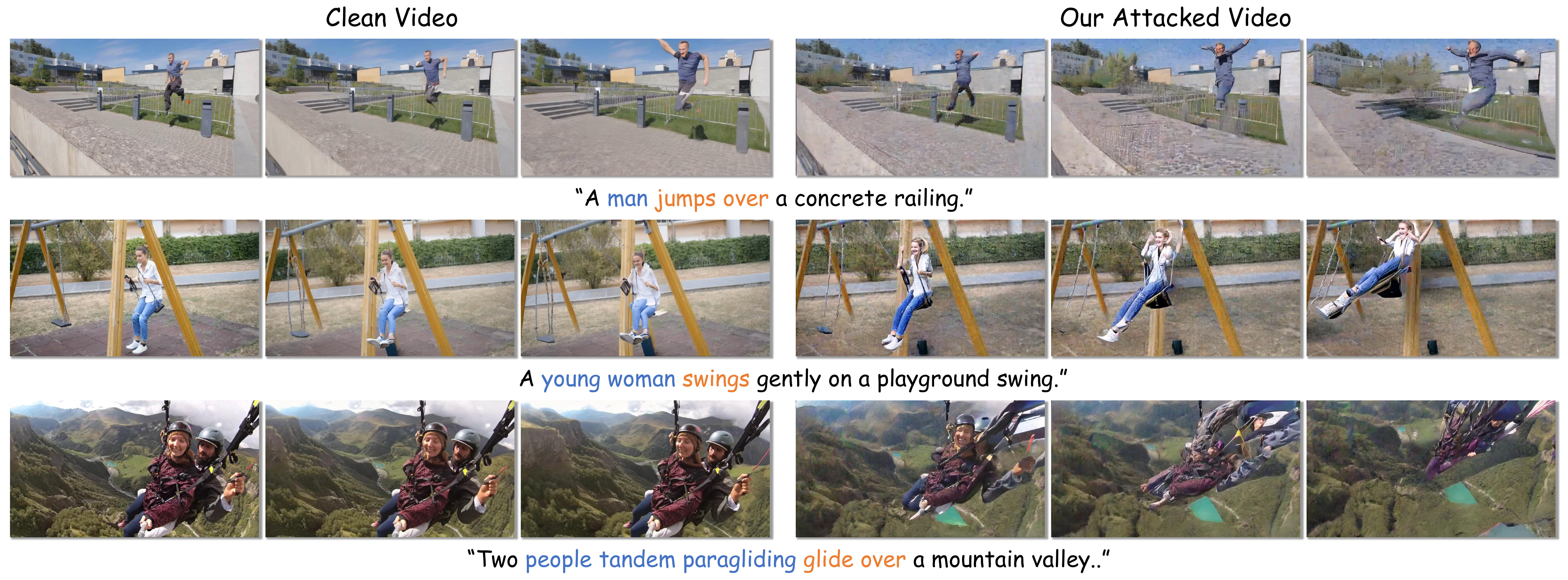}
     \caption{Clean and attacked video results on I2VGen-XL using Immune2V.}
    \label{fig:i2vgenxl}
\end{figure}


We additionally evaluate Immune2V on two earlier dual-stream I2V models, DynamiCrafter \cite{xing2024dynamicrafter} and I2VGen-XL \cite{zhang2023i2vgen}, and provide qualitative examples in Figs.~\ref{fig:dynamicrafter} and~\ref{fig:i2vgenxl}. In both cases, the attacks easily lead to visible distortions and motion inconsistencies, further confirming that our approach generalizes beyond the Wan~2.1 pipeline.

Attacking these earlier models appears \textbf{significantly easier} than attacking Wan~2.1. We conjecture that this is largely due to architectural and training differences, such as older VAE designs and more limited training data. While strong degradation can be readily achieved on these earlier models, this does not necessarily imply that the I2V immunization task itself is trivial. Instead, stronger and more modern models provide a more challenging and informative testbed for systematically analyzing the structure of dual-stream I2V pipelines and designing targeted attack strategies.

\clearpage
\newpage

\section{Challenging Cases}

The effectiveness of the attack mainly depends on the contrast introduced by the conditioning signals, specifically the choice of the target image and the bad prompt. Intuitively, when the target image is sufficiently different from the source image and the bad prompt is strongly unrelated to the original content, the spatial-temporal and semantic objectives introduce stronger deviations in the intermediate representations, which typically lead to more pronounced structural collapse in the generated video. If the target image is visually similar to the source image, it may not introduce sufficient distortion in the latent representation, making the optimization less effective. Similarly, when the bad prompt fails to induce a strong semantic shift, it cannot provide enough conflicting conditioning signals to significantly disrupt the generation process.

The attack effectiveness can also depend on the properties of the input image itself. Scenes with a clear foreground subject and well-defined motion patterns tend to exhibit stronger attack effects, as the generation process relies heavily on the conditioning signals. In contrast, images with weak motion cues or ambiguous semantics may not provide sufficient structural guidance for the generation process, making the attack outcome less predictable.

\newpage
\section{Broader Impact}

\subsection{Positive Societal Impact}

The proposed Immune2V framework aims to improve the safety of image-to-video generation systems by enabling proactive protection of visual content. As generative video models become increasingly capable \cite{xing2024survey, ma2025controllable}, concerns about unauthorized manipulation and synthetic media creation have also grown \cite{hussain2025toward, popa2025deepfake}. By introducing an image immunization mechanism, our method allows creators and platform providers to reduce the risk firsthand that images are later used to generate misleading or manipulated videos. Such protection mechanism is particularly valuable for journalists, public figures, and individuals whose images are frequently shared online. Our work therefore contributes to the growing line of research on defensive technologies that mitigate potential harms of generative media.

\subsection{Potential Misuse and Responsible Release}

Although Immune2V is designed as a defensive mechanism, adversarial perturbation techniques may potentially be repurposed in unintended ways. For example, similar optimization procedures could be adapted to manipulate generative systems or interfere with other machine learning models. As with many security-related research areas, there exists a dual-use risk where defensive insights could inspire new forms of attacks. To mitigate such risks, we frame our work primarily as a protection mechanism for visual content rather than a tool for exploiting generative systems. Our experimental evaluation focuses on understanding vulnerabilities in current I2V pipelines and developing countermeasures against misuse. 

\subsection{Identity and Face-Specific Considerations}

Image-to-video generation systems raise important concerns when applied to identity-related visual content. Synthetic videos generated from personal images may contribute to misinformation, reputational harm, or non-consensual media creation. In this context, protective methods such as Immune2V may serve as a potential safeguard against the unauthorized reuse of visual content in generative pipelines.

Identity preservation in generative video systems involves not only facial images but also broader identity-related visual signals. Distinctive clothing, personal belongings, private environments, or contextual cues may all be used to falsely suggest that a person appeared in a particular place, participated in a certain event, or engaged in an action that never occurred. These risks highlight the importance of considering identity-related information beyond facial appearance when developing protection mechanisms against generative misuse.

At the same time, building reliable safeguards for identity-related content remains a long-term challenge. Future work may further investigate how protection mechanisms can generalize across different forms of personal visual information while maintaining the responsible use of generative technologies.

\stopcontents[appendices]


\end{document}